\documentclass{article}
  
\usepackage{times}
\usepackage{graphicx} % more modern
\usepackage{subfigure} 

\usepackage{natbib}

\usepackage{algorithm}
\usepackage{algorithmic}

\usepackage{hyperref}

\usepackage[accepted]{icml2017}

\usepackage{amsmath,amsfonts,amssymb,bbm,epsfig,bm}
\usepackage{balance}
\usepackage{booktabs}
\usepackage{multirow}
\usepackage{array}

\renewcommand{\algorithmicrequire}{\textbf{Input:}}
\renewcommand{\algorithmicensure}{\textbf{Output:}}

\newcommand{\E}{\mathbb{E}}
\newcommand{\KL}{\text{KL}}

\newcolumntype{L}[1]{>{\raggedright\arraybackslash}p{#1}}
\newcolumntype{R}[1]{>{\raggedleft\arraybackslash}p{#1}}
\newcolumntype{?}{!{\vrule width 1pt}}

\usepackage{color}

\icmltitlerunning{Toward Controlled Generation of Text}

\begin{document} 

\twocolumn[
\icmltitle{Toward Controlled Generation of Text}

\begin{icmlauthorlist}
\icmlauthor{Zhiting Hu}{1,2}
\icmlauthor{Zichao Yang}{1}
\icmlauthor{Xiaodan Liang}{1,2}
\icmlauthor{Ruslan Salakhutdinov}{1}
\icmlauthor{Eric P. Xing}{1,2}
\end{icmlauthorlist}

\icmlaffiliation{1}{Carnegie Mellon University}
\icmlaffiliation{2}{Petuum, Inc.}

\icmlcorrespondingauthor{Zhiting Hu}{zhitingh@cs.cmu.edu}

\icmlkeywords{Deep generative models, text generation, variational autoencoders, wake sleep algorithm, disentangled representation learning}

\vskip 0.3in
]

\printAffiliationsAndNotice{} 

\begin{abstract} 
Generic generation and manipulation of text is challenging and has limited success compared to recent deep generative modeling in visual domain. This paper aims at generating plausible text sentences, whose attributes are controlled by learning disentangled latent representations with designated semantics. We propose a new neural generative model which combines variational auto-encoders (VAEs) and holistic attribute discriminators for effective imposition of semantic structures. The model can alternatively be seen as enhancing VAEs with the wake-sleep algorithm for leveraging fake samples as extra training data. With differentiable approximation to discrete text samples, explicit constraints on independent attribute controls, and efficient collaborative learning of generator and discriminators, our model learns interpretable representations from even only word annotations, and produces sentences with desired attributes of sentiment and tenses. Quantitative experiments using trained classifiers as evaluators validate the accuracy of short sentence and attribute generation.
\end{abstract}

\section{Introduction}
There is a surge of research interest in deep generative models~\cite{hu2017unifying}, such as Variational Autoencoders (VAEs)~\cite{kingma2013auto}, Generative Adversarial Nets (GANs)~\cite{goodfellow2014generative}, and auto-regressive models~\cite{van2016pixel}. Despite their impressive advances in visual domain, such as image generation~\cite{radford2015unsupervised}, learning interpretable image representations~\cite{chen2016infogan}, and image editing~\cite{zhu2016generative}, applications to natural language generation have been relatively less studied. 
Even generating realistic sentences is challenging as the generative models are required to capture complex semantic structures underlying sentences. Previous work have been mostly limited to task-specific applications in supervised settings, including machine translation~\cite{bahdanau2014neural} and image captioning~\cite{vinyals2015show}. However, autoencoder frameworks~\cite{sutskever2014sequence} and recurrent neural network language models~\cite{mikolov2010recurrent} do not apply to generic text generation from arbitrary hidden representations due to the unsmoothness of effective hidden codes~\cite{bowman2015generating}.
Very few recent attempts of using VAEs~\cite{bowman2015generating,tang2016unsupervised} and GANs~\cite{yu2016seqgan,zhang2016generating} have been made to investigate generic text generation, while their generated text is largely randomized and uncontrollable.

In this paper we tackle the problem of {\it controlled} generation of text. That is, we focus on generating realistic sentences, whose attributes can be controlled by learning disentangled latent representations. To enable the manipulation of generated sentences, a few challenges need to be addressed.

A first challenge comes from the discrete nature of text samples.
The resulting non-differentiability hinders the use of global discriminators that assess generated samples and back-propagate gradients to guide the optimization of generators in a holistic manner, as shown to be highly effective in continuous image generation and representation modeling~\cite{chen2016infogan,larsen2016autoencoding,dosovitskiy2016generating}. A number of recent approaches attempt to address the non-differentiability through policy learning~\cite{yu2016seqgan} which tends to suffer from high variance during training, or continuous approximations~\cite{zhang2016generating,kusner2016gans} where only preliminary qualitative results are presented. As an alternative to the discriminator based learning, semi-supervised VAEs~\cite{kingma2014semi} minimize element-wise reconstruction error on observed examples and are applicable to discrete visibles. 
This, however, loses the holistic view of full sentences and can be inferior especially for modeling global abstract attributes (e.g., sentiment). 

Another challenge for controllable generation 
relates to learning disentangled latent representations. Interpretability expects each part of the latent representation to govern and {\it only} focus on one aspect of the samples. Prior methods~\cite{chen2016infogan,odena2016conditional} on structured representation learning lack explicit enforcement of the independence property on the full latent representation, and varying individual code may result in unexpected variation of other unspecified attributes besides the desired one. 

In this paper, we propose a new text generative model that addresses the above issues, permitting highly disentangled representations with designated semantic structure, and generating sentences with dynamically specified attributes. We base our generator on VAEs in combination with holistic discriminators of attributes for effective imposition of structures on the latent code. End-to-end optimization is enabled with differentiable softmax approximation which anneals smoothly to discrete case and helps fast convergence.
The probabilistic encoder of VAE also functions as an additional discriminator to capture variations of implicitly modeled aspects, and guide the generator to avoid entanglement during attribute code manipulation.

Our model can be interpreted as enhancing VAEs with an extended wake-sleep procedure~\cite{hinton1995wake}, where the sleep phase enables incorporation of generated samples for learning both the generator and discriminators in an alternating manner. The generator and the discriminators effectively provide feedback signals to each other, resulting in an efficient mutual bootstrapping framework. 
We show a little supervision (e.g., 100s of annotated sentences) is sufficient to learn structured representations.

Besides efficient representation learning and enabled semi-supervised training, another advantage of using discriminators as learning signals for the generator, as compared to conventional conditional reconstruction based methods~\cite{wen2015semantically,kingma2014semi}, is that discriminators of different attributes can be trained independently. That is, for each attribute one can use separate labeled data for training the respective discriminator, and the trained discriminators can be combined arbitrarily to control a set of attributes of interest. In contrast, reconstruction based approaches typically require every instance of the training data to be labeled exhaustively  with all target attributes~\cite{wen2015semantically}, or to marginalize out any missing attributes~\cite{kingma2014semi} which can be computationally expensive.  

As a showing case, we apply our model to generate sentences with controlled sentiment and tenses. Though to our best knowledge there is no text corpus with both sentiment and tense labels, our method enables to use separate datasets, one with annotated sentiment and the other with tense labels. Quantitative experiments demonstrate the efficacy of our method. Our model improves over previous generative models on the accuracy of generating specified attributes as well as performing classification using generated samples. We show our method learns highly disentangled representations from only word-level labels, and produces plausible short sentences.
\section{Related Work}\label{sec:related}
Remarkable progress has been made in deep generative modeling. \citet{hu2017unifying} provide a unified view of a diverse set of deep generative methods. Variational Autoencoders (VAEs)~\cite{kingma2013auto} consist of encoder and generator networks which encode a data example to a latent representation and generate samples from the latent space, respectively. The model is trained by maximizing a variational lower bound on the data log-likelihood under the generative model. A KL divergence loss is minimized to match the posterior of the latent code with a prior, which enables every latent code from the prior to decode into a plausible sentence. Without the KL regularization, VAEs degenerate to autoencoders and become inapplicable for the generic generation. The vanilla VAEs are incompatible with discrete latents as they hinder differentiable parameterization for learning the encoder. Wake-sleep algorithm~\cite{hinton1995wake} introduced for learning deep directed graphical models shares similarity with VAEs by also combining an inference network with the generator. The wake phase updates the generator with samples generated from the inference network on training data, while the sleep phase updates the inference network based on samples from the generator. Our method combines VAEs with an extended wake-sleep in which the sleep procedure updates both the generator and inference network (discriminators), enabling collaborative semi-supervised learning.
 
Besides reconstruction in raw data space, discriminator-based metric provides a different way for generator learning, i.e., the discriminator assesses generated samples and feedbacks learning signals. For instance, GANs~\cite{goodfellow2014generative} use a discriminator to feedback the probability of a sample being recognized as a real example. \citet{larsen2016autoencoding} combine VAEs with GANs for enhanced image generation. \citet{dosovitskiy2016generating,taigman2017unsupervised} use discriminators to measure high-level perceptual similarity. Applying discriminators to text generation is hard due to the non-differentiability of discrete samples~\cite{yu2016seqgan,zhang2016generating,kusner2016gans}.
\citet{bowman2015generating,tang2016unsupervised,yang2017improved} instead use VAEs without discriminators. All these text generation methods do not learn disentangled latent representations, resulting in randomized and uncontrollable samples. In contrast, disentangled generation in visual domain has made impressive progress. E.g., InfoGAN~\cite{chen2016infogan}, which resembles the extended sleep procedure of our joint VAE/wake-sleep algorithm, disentangles latent representation in an unsupervised manner. The semantic of each dimension is observed after training rather than designated by users in a controlled way. \citet{siddharth2017learning,kingma2014semi} base on VAEs and obtain disentangled image representations with semi-supervised learning. \citet{zhou2017multi} extend semi-supervised VAEs for text transduction. In contrast, our model combines VAEs with discriminators which provide a better, holistic metric compared to element-wise reconstruction.
Moreover, most of these approaches have only focused on the disentanglement of the structured part of latent representations, while ignoring potential dependence of the structured code with attributes not explicitly encoded. We address this by introducing an independency constraint, and show its effectiveness for improved interpretability.
\section{Controlled Generation of Text}\label{sec:model}
Our model aims to generate plausible sentences conditioned on representation vectors which are endowed with designated semantic structures. 
For instance, to control sentence sentiment, our model allocates one dimension of the latent representation to encode ``positive'' and ``negative'' semantics, and generates samples with desired sentiment by simply specifying a particular code.
Benefiting from the disentangled structure, each such code is able to capture a salient attribute and is independent with other features. 
Our deep text generative model possesses several merits compared to prior work, as it 1) facilitates effective imposition of latent code semantics by enabling global discriminators to guide the discrete text generator learning; 2) improves model interpretability by explicitly enforcing the constraints on independent attribute controls; 3) permits efficient semi-supervised learning and bootstrapping by  synthesizing variational auto-encoders with a tailored wake-sleep approach.
We first present the overview of our framework (\S\ref{sec:overview}), then describe the model in detail (\S\ref{sec:struct}). 

\subsection{Model Overview}\label{sec:overview}
We build our framework starting from variational auto-encoders (\S\ref{sec:related}) which have been used for text generation~\cite{bowman2015generating}, where sentence $\hat{\bm{x}}$ is generated conditioned on latent code $\bm{z}$. The vanilla VAE employs an unstructured vector $\bm{z}$ in which the dimensions are entangled.
To model and control the attributes of interest in an interpretable way, we augment the unstructured variables $\bm{z}$ with a set of structured variables $\bm{c}$ each of which targets a salient and independent semantic feature of sentences. 

We want our sentence generator to condition on the combined vector $(\bm{z},\bm{c})$, and generate samples that fulfill the attributes as specified in the structured code $\bm{c}$. 
Conditional generation in the context of VAEs (e.g., semi-supervised VAEs~\cite{kingma2014semi}) is often learned by reconstructing observed examples given their feature code. However, as demonstrated in visual domain, compared to computing element-wise distances in the data space, computing distances in the feature space allows invariance to distracting transformations and provides a better, holistic metric.
Thus, for each attribute code in $\bm{c}$, we set up an individual discriminator to measure how well the generated samples match the desired attributes, and drive the generator to produce improved results. 
The difficulty of applying discriminators in our context is that text samples are discrete and non-differentiable, which breaks down gradient propagation from the discriminators to the generator. We use a continuous approximation based on softmax with a decreasing temperature, which anneals to the discrete case as training proceeds. This simple yet effective approach enjoys low variance and fast convergence.

Intuitively, having an interpretable representation would imply that each structured code in $\bm{c}$ can independently control its target feature, without entangling with other attributes, especially those not explicitly modeled. We encourage the independency by enforcing those irrelevant attributes to be completely captured in the unstructured code $\bm{z}$ and thus be separated from $\bm{c}$ that we will manipulate. 
To this end, we reuse the VAE encoder as an additional discriminator for recognizing the attributes modeled in $\bm{z}$, and train the generator so that these unstructured attributes can be recovered from the generated samples. 
As a result, varying different attribute codes will keep the unstructured attributes invariant as long as $\bm{z}$ is unchanged.

\begin{figure}[t]
\begin{center}
\includegraphics[width=0.85\columnwidth]{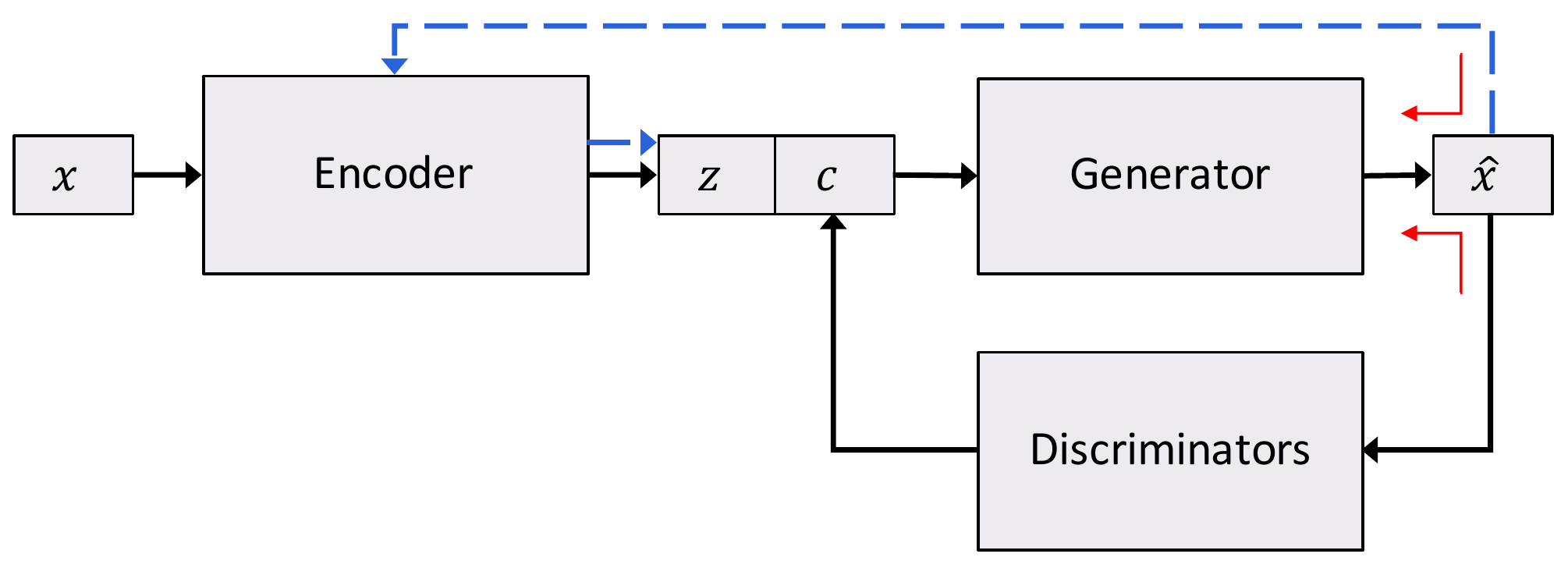}
\vspace{-14pt}
\caption{The generative model, where $\bm{z}$ is unstructured latent code and $\bm{c}$ is structured code targeting sentence attributes to control. Blue dashed arrows denote the proposed independency constraint (section~\ref{sec:struct} for details), and red arrows denote gradient propagation enabled by the differentiable approximation.}
\label{fig:arch}
\end{center}
\vspace{-12pt}
\end{figure}

Figure~\ref{fig:arch} shows the overall model structure. Our complete model incorporates VAEs and attribute discriminators, in which the VAE component trains the generator to reconstruct real sentences for generating plausible text, while the discriminators enforce the generator to produce attributes coherent with the conditioned code. The attribute discriminators are learned to fit labeled examples to entail designated semantics, as well as trained to explain samples from the generator. That is, the generator and the discriminators form a pair of collaborative learners and provide feedback signals to each other. The collaborative optimization resembles wake-sleep algorithm. We show the combined VAE/wake-sleep learning enables a highly efficient semi-supervised framework, which requires only a little supervision to obtain interpretable representation and generation.

\subsection{Model Structure}\label{sec:struct}
We now describe our model in detail, by presenting the learning of generator and discriminators, respectively. 

\subparagraph{Generator Learning}\quad\\
The generator $G$ is an LSTM-RNN for generating token sequence $\hat{\bm{x}}=\{\hat{x}_1,\dots,\hat{x}_T\}$ conditioned on the latent code $(\bm{z},\bm{c})$, which depicts a generative distribution:
\begin{equation}
\small
\begin{split}
\hat{\bm{x}} \sim G(\bm{z},\bm{c}) &= p_{G}(\hat{\bm{x}}|\bm{z},\bm{c}) \\  
&=\prod\nolimits_{t}p(\hat{x}_{t}|\hat{\bm{x}}^{<t},\bm{z},\bm{c}),
\end{split}
\label{eq:gen_x}
\end{equation}
where $\hat{\bm{x}}^{<t}$ indicates the tokens preceding $\hat{x}_{t}$.
The generation thus involves a sequence of discrete decision making which samples a token from a multinomial distribution parametrized using softmax function at each time step $t$:
\begin{equation}
\small
\begin{split}
\hat{x}_t \sim \text{softmax}(\bm{o}_t / \tau),
\end{split}
\label{eq:softmax}
\end{equation}
where $\bm{o}_t$ is the logit vector as the inputs to the softmax function, and $\tau>0$ is the temperature normally set to $1$.

The unstructured part $\bm{z}$ of the representation is modeled as continuous variables with standard Gaussian prior $p(\bm{z})$, while the structured code $\bm{c}$ can contain both continuous and discrete variables to encode different attributes (e.g., sentiment categories, formality) with appropriate prior $p(\bm{c})$. Given observation $\bm{x}$, the base VAE includes a conditional probabilistic encoder $E$ to infer the latents $\bm{z}$:
\begin{equation}
\small
\begin{split}
\bm{z} \sim E(\bm{x}) = q_E(\bm{z} | \bm{x}).
\end{split}
\end{equation}
Let $\bm{\theta}_G$ and $\bm{\theta}_E$ denote the parameters of the generator $G$ and the encoder $E$, respectively.
The VAE is then optimized to minimize the reconstruction error of observed real sentences, and at the same time regularize the encoder to be close to the prior $p(\bm{z})$:
\begin{equation}
\small
\begin{split}
\mathcal{L}_{\text{VAE}}(\bm{\theta}_G, \bm{\theta}_E; \bm{x}) =\  &\KL(q_{E}(\bm{z}|\bm{x}) \| p(\bm{z})) \\ 
&- \E_{q_E(\bm{z}|\bm{x})q_D(\bm{c}|\bm{x})}\left[\log p_{G}(\bm{x}|\bm{z},\bm{c}) \right],
\end{split}
\label{eq:loss-vae}
\end{equation}
where $\KL(\cdot\|\cdot)$ is the KL-divergence; and $q_{D}(\bm{c}|\bm{x})$ is the conditional distribution defined by the discriminator $D$ for each structured variable in $\bm{c}$:
\begin{equation}
\small
\begin{split}
D(\bm{x}) = q_{D}(\bm{c}|\bm{x}).
\end{split}
\end{equation}
Here, for notational simplicity, we assume only one structured variable and thus one discriminator, though our model specification can straightforwardly be applied to many attributes. The distribution over $(\bm{z}, \bm{c})$ factors into $q_E$ and $q_D$ as we are learning disentangled representations. Note that here the discriminator $D$ and code $\bm{c}$ are not learned with the VAE loss, but instead optimized with the objectives described shortly. Besides the reconstruction loss which drives the generator to produce realistic sentences, the discriminator provides extra learning signals which enforce the generator to produce coherent attribute that matches the structured code in $\bm{c}$. 
However, as it is impossible to propagate gradients from the discriminator through the discrete samples, we resort to a deterministic continuous approximation. The approximation replaces the sampled token $\hat{x}_t$ (represented as a one-hot vector) at each step with the probability vector in Eq.\eqref{eq:softmax} which is differentiable w.r.t the generator's parameters. The probability vector is used as the output at the current step and the input to the next step along the sequence of decision making. The resulting ``soft'' generated sentence, denoted as $\widetilde{G}_\tau(\bm{z},\bm{c})$, is fed into the discriminator\footnote{The probability vector thus functions to average over the word embedding matrix to obtain a ``soft'' word embedding at each step.} to measure the fitness to the target attribute, leading to the following loss for improving $G$:
\begin{equation}
\small
\begin{split}
\mathcal{L}_{\text{Attr},c}(\bm{\theta}_G) = -\E_{p(\bm{z})p(\bm{c})}\left[ \log q_{D}(\bm{c} | \widetilde{G}_\tau(\bm{z},\bm{c})) \right].
\end{split}
\label{eq:loss-attr}
\end{equation}
The temperature $\tau$ (Eq.\ref{eq:softmax}) is set to $\tau\to 0$ as training proceeds, yielding increasingly peaked distributions that finally emulate discrete case. The simple deterministic approximation effectively leads to reduced variance and fast convergence during training, which enables efficient learning of the conditional generator. The diversity of generation results is guaranteed since we use the approximation only for attribute modeling and the base sentence generation is learned through VAEs.

With the objective in Eq.\eqref{eq:loss-attr}, each structured attribute of generated sentences is controlled through the corresponding code in $\bm{c}$ and is independent with other variables in the latent representation.
However, it is still possible that other attributes not explicitly modeled may also entangle with the code in $\bm{c}$, and thus varying a dimension of $\bm{c}$ can yield unexpected variation of these attributes we are not interested in. To address this, we introduce the independency constraint which separates these attributes with $\bm{c}$ by enforcing them to be fully captured by the unstructured part $\bm{z}$.
Therefore, besides the attributes explicitly encoded in $\bm{c}$, we also train the generator so that other non-explicit attributes can be correctly recognized from the generated samples and match the unstructured code $\bm{z}$.
Instead of building a new discriminator, we reuse the variational encoder $E$ which serves precisely to infer the latents $\bm{z}$ in the base VAE. The loss is in the same form as with Eq.\eqref{eq:loss-attr} except replacing the discriminator conditional $q_D$ with the encoder conditional~$q_E$:
\begin{equation}
\small
\begin{split}
\mathcal{L}_{\text{Attr},z}(\bm{\theta}_G) = -\E_{p(\bm{z})p(\bm{c})}\left[ \log q_{E}(\bm{z} | \widetilde{G}_\tau(\bm{z},\bm{c})) \right].
\end{split}
\label{eq:loss-attr-z}
\end{equation}
Note that, as the discriminator in Eq.(\ref{eq:loss-attr}), the encoder now performs inference over generated samples from the prior, as opposed to observed examples as in VAEs.

Combining Eqs.\eqref{eq:loss-vae}-\eqref{eq:loss-attr-z} we obtain the generator objective:
\begin{equation}
\small
\begin{split}
\min\nolimits_{\bm{\theta}_G} \mathcal{L}_{G} = \mathcal{L}_{\text{VAE}} + \lambda_c \mathcal{L}_{\text{Attr},c} + \lambda_z \mathcal{L}_{\text{Attr},z},
\end{split}
\label{eq:loss-g}
\end{equation}
where $\lambda_c$ and $\lambda_z$ are balancing parameters. The variational encoder is trained by minimizing the VAE loss, i.e., $\min_{\bm{\theta}_E} \mathcal{L}_{\text{VAE}}$.

\paragraph{Discriminator Learning}\quad\\
The discriminator $D$ is trained to accurately infer the sentence attribute and evaluate the error of recovering the desired feature as specified in the latent code. For instance, for categorical attribute, the discriminator can be formulated as a sentence classifier; while for continuous target a probabilistic regressor can be used. 
The discriminator is learned in a different way compared to the VAE encoder, since the target attributes can be discrete which are not supported in the VAE framework. 
Moreover, in contrast to the unstructured code $\bm{z}$ which is learned in an unsupervised manner, the structured variable $\bm{c}$ uses labeled examples to entail designated semantics. We derive an efficient semi-supervised learning method for the discriminator. 

Formally, let $\bm{\theta}_D$ denote the parameters of the discriminator. To learn specified semantic meaning, we use a set of labeled examples $\mathcal{X}_L = \{(\bm{x}_L, \bm{c}_L)\}$ to train the discriminator $D$ with the following objective:
\begin{equation}
\small
\begin{split}
\mathcal{L}_{s}(\bm{\theta}_D) = - \E_{\mathcal{X}_L}\left[ \log q_{D}(\bm{c}_L | \bm{x}_L) \right].
\end{split}
\end{equation}
Besides, the conditional generator $G$ is also capable of synthesizing (noisy) sentence-attribute pairs $(\hat{\bm{x}}, \bm{c})$ which can be used to augment training data for semi-supervised learning. To alleviate the issue of noisy data and ensure robustness of model optimization, we incorporate a minimum entropy regularization term~\cite{grandvalet2004semi,reed2014training}. The resulting objective is thus:
\begin{equation}
\small
\begin{split}
\mathcal{L}_{u}(\bm{\theta}_D) = - \E_{p_{G}(\hat{\bm{x}}|\bm{z},\bm{c})p(\bm{z})p(\bm{c})}\big[ \log q_{D}(\bm{c}|\hat{\bm{x}}) + \beta \mathcal{H}(q_{D}(\bm{c}'|\hat{\bm{x}})) \big],
\end{split}
\label{eq:loss-d-u}
\end{equation}
where $\mathcal{H}(q_{D}(\bm{c}'|\hat{\bm{x}}))$ is the empirical Shannon entropy of distribution $q_D$ evaluated on the generated sentence $\hat{\bm{x}}$; and $\beta$ is the balancing parameter. 
Intuitively, the minimum entropy regularization encourages the model to have high confidence in predicting labels. 

The joint training objective of the discriminator using both labeled examples and synthesized samples is then given as:
\begin{equation}
\small
\begin{split}
\min\nolimits_{\bm{\theta}_D} \mathcal{L}_{D} = \mathcal{L}_s + \lambda_u \mathcal{L}_u,
\end{split}
\label{eq:loss-d}
\end{equation}
where $\lambda_u$ is the balancing parameter. 

% algorithm
\begin{algorithm}[t]
\small
\centering
\caption{\small Controlled Generation of Text}
\label{alg:opt}
\begin{algorithmic}[1]
\REQUIRE A large corpus of unlabeled sentences $\mathcal{X}=\{\bm{x}\}$ \\
\quad\ \  A few sentence attribute labels $\mathcal{X}_L = \{(\bm{x}_L, \bm{c}_L)\}$ \\
\quad\ \  Parameters: $\lambda_c, \lambda_z, \lambda_u, \beta$  -- balancing parameters \\
\STATE Initialize the base VAE by minimizing Eq.\eqref{eq:loss-vae} on $\mathcal{X}$ with $\bm{c}$ sampled from prior $p(\bm{c})$
\REPEAT
    \STATE Train the discriminator $D$ by Eq.\eqref{eq:loss-d}
    \STATE Train the generator $G$ and the encoder $E$ by Eq.\eqref{eq:loss-g} and minimizing Eq.\eqref{eq:loss-vae}, respectively.
\UNTIL{convergence}
\ENSURE Sentence generator $G$ conditioned on disentangled representation $(\bm{z},\bm{c})$
\end{algorithmic}
\end{algorithm}

\paragraph{Summarization and Discussion}\quad\\
We have derived our model and its learning procedure. The generator is first initialized by training the base VAE on a large corpus of unlabeled sentences, through the objective of minimizing Eq.\eqref{eq:loss-vae} with the latent code $\bm{c}$ at this time sampled from the prior distribution $p(\bm{c})$. 
The full model is then trained by alternating the optimization of the generator and the discriminator, as summarized in Algorithm~\ref{alg:opt}. 

\begin{figure}[t]
\begin{center}
\includegraphics[width=0.85\columnwidth]{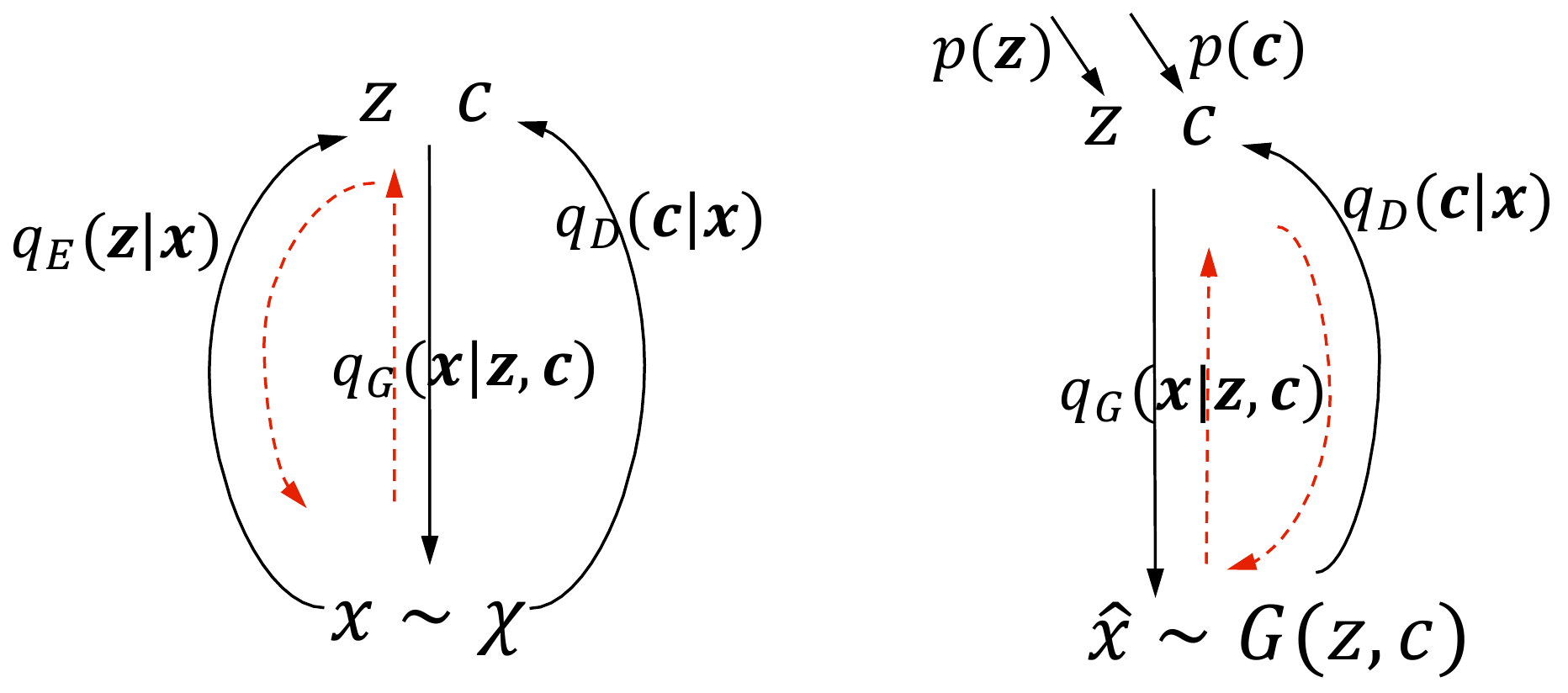}
\vspace{-12pt}
\caption{{\bf Left:} The VAE and wake procedure, corresponding to Eq.\eqref{eq:loss-vae}. {\bf Right:} The sleep procedure, corresponding to Eqs.\eqref{eq:loss-attr}-\eqref{eq:loss-attr-z} and \eqref{eq:loss-d-u}. Black arrows denote inference and generation; red dashed arrows denote gradient propagation. The two steps in the sleep procedure, i.e., optimizing the discriminator and the generator, respectively, are performed in an alternating manner.}
\label{fig:wake-sleep}
\end{center}
\vspace{-15pt}
\end{figure}

Our model can be viewed as combining the VAE framework with an extended wake-sleep method, as illustrated in Figure~\ref{fig:wake-sleep}. Specifically, in Eq.\eqref{eq:loss-d-u}, samples are produced by the generator and used as targets for maximum likelihood training of the discriminator. This resembles the sleep phase of wake-sleep. Eqs.\eqref{eq:loss-attr}-\eqref{eq:loss-attr-z} further leverage the generated samples to improve the generator. We can see the above together as an extended sleep procedure based on ``dream'' samples obtained by ancestral sampling from the generative network. On the other hand, Eq.\eqref{eq:loss-vae} samples $\bm{c}$ from the discriminator distribution $q_D(\bm{c}|\bm{x})$ on observation $\bm{x}$, to form a target for training the generator, which corresponds to the wake phase. The effective combination enables discrete latent code, holistic discriminator metrics, and efficient mutual bootstrapping.

Training of the discriminators need supervised data to impose designated semantics. Discriminators for different attributes can be trained independently on separate labeled sets. That is, the model does not require a sentence to be annotated with all attributes, but instead needs only independent labeled data for each individual attribute. Moreover, as the labeled data are used only for learning attribute semantics instead of direct sentence generation, we are allowed to extend the data scope beyond labeled sentences to, e.g., labeled words or phrases. As shown in the experiments (section~\ref{sec:exp}), our method is able to effectively lift the word level knowledge to sentence level and generate convincing sentences. Finally, with the augmented unsupervised training in the sleep phrase, we show a little supervision is sufficient for learning structured representations.
\section{Experiments}\label{sec:exp}
We apply our model to generate short sentences (length $\leq$ 15) with controlled sentiment and tense. Quantitative experiments using trained classifiers as evaluators show our model gives improved generation accuracy. 
Disentangled representation is learned with a few labels or only word annotations. We also validate the effect of the proposed independency constraint for interpretable generation.
%
%\subsection{Setup}
\paragraph{Datasets}\quad\\
{\bf Sentence corpus. } We use a large IMDB text corpus~\cite{diao2014jointly} for training the generative models. This is a collection of 350K movie reviews. We select sentences containing at most $15$ words, and replace infrequent words with the token ``$<$unk$>$''. The resulting dataset contains around 1.4M sentences with the vocabulary size of 16K.

{\bf Sentiment. } To control the sentiment (``positive'' or ``negative'') of generated sentences, we test on the following labeled sentiment data: (1) Stanford Sentiment Treebank-2 ({\bf SST-full})~\cite{socher2013recursive} consists of 6920/872/1821 movie review sentences with binary sentiment annotations in the train/dev/test sets, respectively. We use the 2837 training examples with sentence length $\le 15$, and evaluate classification accuracy on the original test set. (2) {\bf SST-small. } To study the size of labeled data required in the semi-supervised learning for accurate attribute control, we sample a small subset from SST-full, containing only 250 labeled sentences for training.
(3) {\bf Lexicon. } We also investigate the effectiveness of our model in terms of using word-level labels for sentence-level control. The lexicon from~\cite{wilson2005recognizing} contains 2700 words with sentiment labels. We use the lexicon for training by treating the words as sentences, and evaluate on the SST-full test set. (4) {\bf IMDB. } We collect a dataset from the IMDB corpus by randomly selecting positive and negative movie reviews. The dataset has 5K/1K/10K sentences in train/dev/test. 

{\bf Tense. }
The second attribute is the tense of the main verb in a sentence.
Though no corpus with sentence tense annotations is readily available, our method is able to learn from only labeled words and generate desired sentences. We compile from the TimeBank (timeml.org) dataset and obtain a lexicon of 5250 words and phrases labeled with one of \{``past'', ``present'', ``future''\}. The lexicon mainly consists of verbs in different tenses (e.g., ``was'', ``will be'') as well as time expressions (e.g., ``in the future'').

Note that our method requires only separate labeled copora for each attribute. And for the tense attribute only annotated words/phrases are used.

%\vspace{-5pt}
\paragraph{Parameter Setting}\quad\\
The generator and encoder are set as single-layer LSTM RNNs with input/hidden dimension of 300 and max sample length of 15. Discriminators are set as ConvNets. Detailed configurations are in the supplements. To avoid vanishingly small KL term in the VAE module (Eq.\ref{eq:loss-vae})~\cite{bowman2015generating}, we use a KL term weight linearly annealing from 0 to 1 during training.  
Balancing parameters are set to $\lambda_c=\lambda_z=\lambda_u=0.1$, and $\beta$ is selected on the dev sets. 
At test time sentences are generated with Eq.\eqref{eq:gen_x}.

\begin{table}[t]
\centering
%\scriptsize 
\small
\begin{tabular}{L{1.5cm} r r r}
\cmidrule[\heavyrulewidth]{1-4}
\multirow{2}{*}{Model}  & \multicolumn{3}{c}{Dataset} \\ \cmidrule(l){2-4}
 & SST-full & SST-small & Lexicon \\ 
\cmidrule[\heavyrulewidth]{1-4}
S-VAE & 0.822 & 0.679 & 0.660 \\
Ours & {\bf 0.851} & {\bf 0.707} & {\bf 0.701} \\
\cmidrule[\heavyrulewidth]{1-4}
\end{tabular}
\vspace{-10pt}
\caption{Sentiment accuracy of generated sentences. S-VAE \cite{kingma2014semi} and our model are trained on the three sentiment datasets and generate 30K sentences, respectively.}
\label{tab:senti}
%\vspace{-15pt}
\end{table}
\subsection{Accuracy of Generated Attributes}
We quantitatively measure sentence attribute control by evaluating the accuracy of generating designated sentiment, and the effect of using samples for training classifiers. We compare with semi-supervised VAE (S-VAE)~\cite{kingma2014semi}, one of the few existing deep models capable of conditional text generation. S-VAE learns to reconstruct observed sentences given attribute code, and no discriminators are used. See \S\ref{sec:related} and~\ref{sec:overview} for more discussions.

We use a state-of-the-art sentiment classifier~\cite{hu2016harnessing} which achieves 90\% accuracy on the SST test set, to automatically evaluate the sentiment generation accuracy. Specifically, we generate sentences given sentiment code $\bm{c}$, and use the pre-trained sentiment classifier to assign sentiment labels to the generated sentences. 
The accuracy is calculated as the percentage of the predictions that match the sentiment code $\bm{c}$.
Table~\ref{tab:senti} shows the results on 30K sentences by the two models which are trained with SST-full, SST-small, and Lexicon, respectively. We see that our method consistently outperforms S-VAE on all datasets. In particular, trained with only 250 labeled examples in SST-small, our model achieves reasonable generation accuracy, demonstrating the ability of learning disentangled representations with very little supervision. More importantly, given only word-level annotations in Lexicon, our model successfully transfers the knowledge to sentence level and generates desired sentiments reasonably well. Compared to our method that drives learning by directly assessing generated sentences, S-VAE attempts to capture sentiment semantics only by reconstructing labeled words, which is less efficient and gives inferior performance.

\begin{figure}[!t]
\begin{center}
\includegraphics[width=0.85\columnwidth]{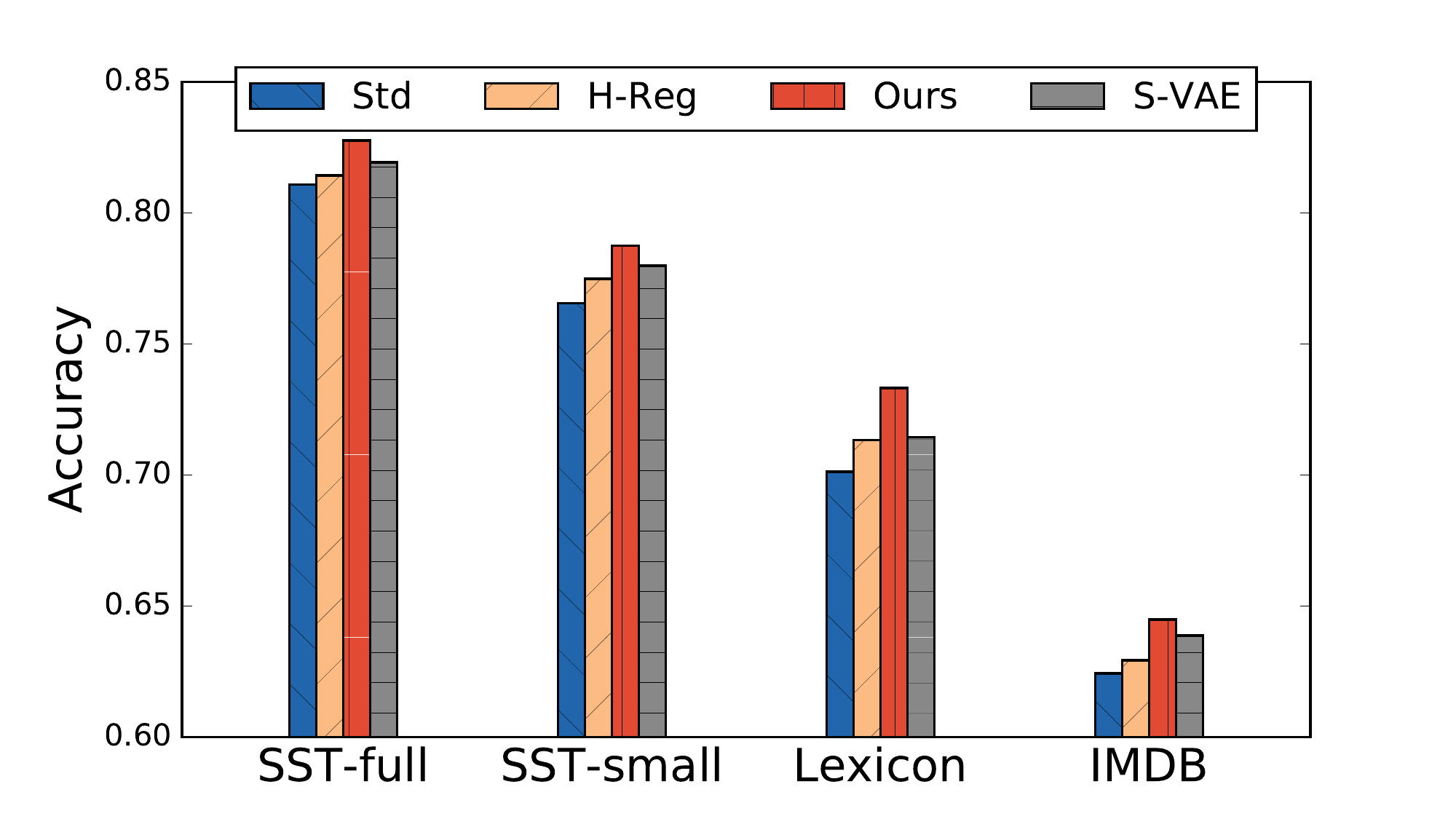}
\vspace{-15pt}
\caption{Test-set accuracy of classifiers trained on four sentiment datasets augmented with different methods (see text for details). The first three datasets use the SST-full test set for evaluation.}
\label{fig:senti-cls}
\end{center}
\vspace{-15pt}
\end{figure}

We next use the generated samples to augment the sentiment datasets and train sentiment classifiers. While not aiming to build best-performing classifiers on these datasets, the classification accuracy serves as an auxiliary measure of the sentence generation quality. That is, higher-quality sentences with more accurate sentiment attribute can predictably help yield stronger sentiment classifiers.
Figure~\ref{fig:senti-cls} shows the accuracy of classifiers trained on the four datasets with different augmentations. ``Std" is a ConvNet trained on the standard original datasets, with the same network structure as with the sentiment discriminator in our model. ``H-reg" additionally imposes the minimum entropy regularization on the generated sentences. ``Ours" incorporates the minimum entropy regularization and the sentiment attribute code $\bm{c}$ of the generated sentences, as in Eq.\eqref{eq:loss-d-u}. S-VAE uses the same protocol as our method to augment with the data generated by the S-VAE model. 
Comparison in Figure~\ref{fig:senti-cls} shows that our method consistently gives the best performance on four datasets. For instance, on Lexicon, our approach achieves 0.733 accuracy, compared to 0.701 of ``Std". The improvement of ``H-Reg" over ``Std" shows positive effect of the minimum entropy regularization on generated sentences. Further incorporating the conditioned sentiment code of the generated samples, as in ``Ours" and ``S-VAE", provides additional performance gains, indicating the advantages of conditional generation for automatic creation of labeled data. Consistent with the above experiment, our model outperforms S-VAE.
\vspace{-10pt}
\subsection{Disentangled Representation}
We study the interpretability of generation and the explicit independency constraint (Eq.\ref{eq:loss-attr-z}) for disentangled control. 

Table~\ref{tab:rep-z-reg} compares the samples generated by models with and without the constraint term, respectively. In the left column where the constraint applies, each pair of sentences, conditioned on different sentiment codes, are highly relevant in terms of, e.g., subject, tone, and wording which are not explicitly modeled in the structured code $\bm{c}$ while instead implicitly encoded in the unstructured code $\bm{z}$. Varying the sentiment code precisely changes the sentiment of the sentences (and paraphrases slightly to ensure fluency), while keeping other aspects unchanged. In contrast, the results in the right column, where the independency constraint is unactivated, show that varying the sentiment code not only changes the polarity of samples, but can also change other aspects unexpected to control, making the generation results less interpretable and predictable.

We demonstrate the power of learned disentangled representation by varying one attribute variable at a time. Table~\ref{tab:rep-tense-senti} shows the generation results. We see that each attribute variable in our model successfully controls its corresponding attribute, and is disentangled with other attribute code. The right column of the table shows meaningful variation of sentence tense as the tense code varies. Note that the semantic of tense is learned only from a lexicon without complete sentence examples. Our model successfully captures the key ingredients (e.g., verb ``was'' for past tense and ``will be'' for future tense) and combines with the knowledge of well-formed sentences to generate realistic samples with specified tense attributes.
Table~\ref{tab:rep-vary-z} further shows generated sentences  with varying code $\bm{z}$ in different settings of structured attribute factors. We obtain samples that are diverse in content while consistent in sentiment and tense. 

We also occasionally observed failure cases as in Table~\ref{tab:failures}, such as implausible sentences, unexpected variations of irrelevant attributes, and inaccurate attribute generations. Improved modeling is expected such as using dilated convolutions as decoder, and decoding with beam search, etc. Better systematic quantitative evaluations are also desired.

\begin{table*}[!h]
\centering
%\scriptsize 
%\small
\begin{tabular}{l l}
\cmidrule[\heavyrulewidth]{1-2}
 {\bf w/ independency constraint} & {\bf w/o independency constraint} \\ \midrule
 the film is strictly routine ! & the acting is bad . \\ 
 the film is full of imagination . & the movie is so much fun . \\
 \\
 after watching this movie , i felt that disappointed . & none of this is very original .   \\
 after seeing this film , i 'm a fan . & highly recommended viewing for its courage , and ideas .  \\
 \\
 the acting is uniformly bad either . &  too bland  \\
 the performances are uniformly good . & highly watchable \\
\\
 this is just awful . & i can analyze this movie without more than three words . \\
 this is pure genius . & i highly recommend this film to anyone who appreciates music .  \\
\cmidrule[\heavyrulewidth]{1-2}
\end{tabular}
\vspace{-4pt}
\caption{Samples from models with or without independency constraint on attribute control (i.e., Eq.\ref{eq:loss-attr-z}). Each pair of sentences are generated with sentiment code set to ``negative'' and ``positive'', respectively, while fixing the unstructured code $\bm{z}$. The SST-full dataset is used for learning the sentiment representation.}
\label{tab:rep-z-reg}
\end{table*}
\begin{table*}[!h]
\centering
%\scriptsize 
%\small
\begin{tabular}{L{7.5cm} l}
\cmidrule[\heavyrulewidth]{1-2}
 {\bf Varying the code of tense} & \\ \midrule
 i thought the movie was too bland and too much  & this was one of the outstanding thrillers of the last decade \\ 
 i guess the movie is too bland and too much & this is one of the outstanding thrillers of the all time \\
 i guess the film will have been too bland & this will be one of the great thrillers of the all time \\
\cmidrule[\heavyrulewidth]{1-2}
\end{tabular}
\vspace{-4pt}
\caption{Each triple of sentences is generated by varying the tense code while fixing the sentiment code and $\bm{z}$.}
\label{tab:rep-tense-senti}
\end{table*}
\begin{table*}[!h]
\centering
%\scriptsize 
%\small
\begin{tabular}{l l}
\cmidrule[\heavyrulewidth]{1-2}
 {\bf Varying the unstructured code $\bm{z}$} &  \\ \midrule
 {\it (``negative'', ``past'')} & {\it (``positive'', ``past'')} \\
 the acting was also kind of hit or miss . & his acting was impeccable \\
 i wish i 'd never seen it & this was spectacular , i saw it in theaters twice \\
 by the end i was so lost i just did n't care anymore & it was a lot of fun \\
 & \\
 {\it (``negative'', ``present'')} & {\it (``positive'', ``present'')} \\
 the movie is very close to the show in plot and characters & this is one of the better dance films \\
 the era seems impossibly distant & i 've always been a big fan of the smart dialogue . \\
 i think by the end of the film , it has confused itself & i recommend you go see this, especially if you hurt \\
 \\
 {\it (``negative'', ``future'')} & {\it (``positive'', ``future'')} \\
 i wo n't watch the movie & i hope he 'll make more movies in the future \\
 and that would be devastating ! &  i will definitely be buying this on dvd \\
 i wo n't get into the story because there really is n't one & you will be thinking about it afterwards, i promise you \\
\cmidrule[\heavyrulewidth]{1-2}
\end{tabular}
\vspace{-4pt}
\caption{Samples by varying the unstructured code $\bm{z}$ given sentiment (``positive''/``negative'') and tense (``past''/``present''/``future'') code.}
\label{tab:rep-vary-z}
\end{table*}
\begin{table*}[!h]
\centering
%\scriptsize 
%\small
\begin{tabular}{L{7.5cm} l}
\cmidrule[\heavyrulewidth]{1-2}
 {\bf Failure cases} & \\ \midrule
 the plot is not so original  &  it does n't get any better the other dance movies \\ 
 the plot weaves us into $<$unk$>$ & it does n't reach them , but the stories look \\
 \\
 he is a horrible actor 's most part & i just think so\\
 he 's a better actor than a standup & i just think !\\
\cmidrule[\heavyrulewidth]{1-2}
\end{tabular}
\vspace{-4pt}
\caption{Failure cases when varying sentiment code with other codes fixed.}
\label{tab:failures}
\end{table*}
\section{Discussions}\label{sec:conclude}
We have proposed a deep generative model that learns interpretable latent representations and generates sentences with specified attributes. We obtained meaningful generation with restricted sentence length, and improved accuracy on sentiment and tense attributes. In the future we would like to improve the modeling and training as above, and extend to generate longer sentences/paragraphs and control more attributes with fine-grained structures.

Our approach combines VAEs with attribute discriminators and imposes explicit independency constraints on attribute controls, enabling disentangled latent code. Semi-supervised learning within the joint VAE/wake-sleep framework is effective with little or incomplete supervision. \citet{hu2017unifying} develop a unified view of a diverse set of deep generative paradigms, including GANs, VAEs, and wake-sleep algorithm. Our model can be alternatively motivated under the view as enhancing VAEs with the extended sleep phase and by leveraging generated samples.

Interpretability of the latent representations not only allows dynamic control of generated attributes, but also provides an interface that connects the end-to-end neural model with conventional structured methods. For instance, we can encode structured constraints (e.g., logic rules or probabilistic structured models) on the interpretable latent code, to incorporate prior knowledge or human intentions~\cite{hu2016harnessing,hu2016deep}; or plug the disentangled generation model into dialog systems to generate natural language responses from structured dialog states~\cite{young2013pomdp}.

Though we have focused on the generation capacity of our model, the proposed collaborative semi-supervised learning framework also helps improve the discriminators by generating labeled samples for data augmentation (e.g., see Figure~\ref{fig:senti-cls}). More generally, for any discriminative task, we can build a conditional generative model to synthesize additional labeled data. The accurate attribute generation of our approach can offer larger performance gains compared to previous generative methods.

{\small
\paragraph{Implementation} We have released code for an adapted version of the proposed algorithm at: \\
\url{https://github.com/asyml/texar/tree/master/examples/text_style_transfer}.

The implementation is based on {\it Texar}~\cite{hu2018texar}, a general-purpose text generation toolkit.
}
%\vspace{-8pt}
{\small 
\paragraph{Acknowledgments} This research is supported by NSF IIS1447676, ONR N000141410684, and ONR N000141712463.
}

%\clearpage
%\balance
\bibliography{icml2017}
\bibliographystyle{icml2017}

\end{document}